\documentclass[11pt]{article}

\usepackage[final]{acl}

\usepackage{times}
\usepackage{latexsym}
\usepackage{algorithm}
\usepackage{algpseudocode}
\usepackage{amsmath}
\usepackage{multirow}
\usepackage{array}
\usepackage{arydshln}

\usepackage[T1]{fontenc}
\usepackage[utf8]{inputenc}

\usepackage{microtype}

\usepackage{inconsolata}
\usepackage[most]{tcolorbox}
\newtcbox{\slotbox}{on line, colback=gray!10, colframe=gray!80, boxrule=0.5pt, arc=2pt, top=1pt, bottom=1pt, left=2pt, right=2pt}

\usepackage{graphicx}
\definecolor{markda}{HTML}{a92b7a}

\definecolor{DarkGreen}{rgb}{0.1,0.5,0.1}

\newcommand{\redacted}{\begin{small}
    \textsf{[REDACTED]}
\end{small}}
\newcommand{\redactedsmall}{\begin{scriptsize}
    \textsf{[REDACTED]}
\end{scriptsize}}

\title{Protecting De-identified Documents from Search-based Linkage Attacks}

\author{Pierre Lison \and Mark Anderson \\
Norwegian Computing Center \\
Oslo, Norway \\
\texttt{\{plison,anderson\}@nr.no} }

\begin{document}
    \maketitle
    \begin{abstract}
        While de-identification models can help conceal the identity of the individual(s)
        mentioned in a document, they fail to address \textit{linkage risks}, defined
        as the potential to map the de-identified text back to its source.  Linkages can potentially propagate harmful outcomes back to individuals and should therefore be prevented. 
        
        One straightforward way to perform such linkages is to extract phrases from
        the de-identified document and then check their presence in the original
        dataset. This paper presents a method %\footnote{The code will be published on GitHub after peer-review.} 
        to counter search-based linkage
        attacks while preserving the semantic integrity of the text. The method
        proceeds in two steps. We first construct an \textit{inverted index} of
        the N-grams occurring in the text collection, making it possible to efficiently
        determine which N-grams appear in less than $k$ documents (either alone
        or in combination with other N-grams). An LLM-based rewriter is then
        iteratively queried to reformulate those text spans until linkage is no
        longer possible. 
        
        Experimental results on two datasets (respectively comprising court cases and Wikipedia biographies) show that the rewriting method can effectively prevent search-based linkages while remaining faithful
        to the original content. However, we also highlight that linkages remain feasible with the help of more advanced, semantics-oriented
        approaches.
    \end{abstract}

    \section{Introduction}

    Most text documents contain personal information in one form or another. To
    alleviate privacy concerns and abide to data protection regulations, many types
    of documents, such as administrative decisions, court cases or patient records,
    must be \textit{de-identified} prior to transferring them to third parties or
    releasing them publicly. De-identification techniques typically operate by
    detecting text spans that express \textit{personally identifiable
    information} (PII) and subsequently masking those from the document \cite{Neamatullah2008-pu,Dernoncourt2017-tk,Lison2021-bf,Szawerna2024-jh,Staab2024-ju,Deusser2025-fs}.

    \begin{figure}[t]
        \includegraphics[scale=0.23]{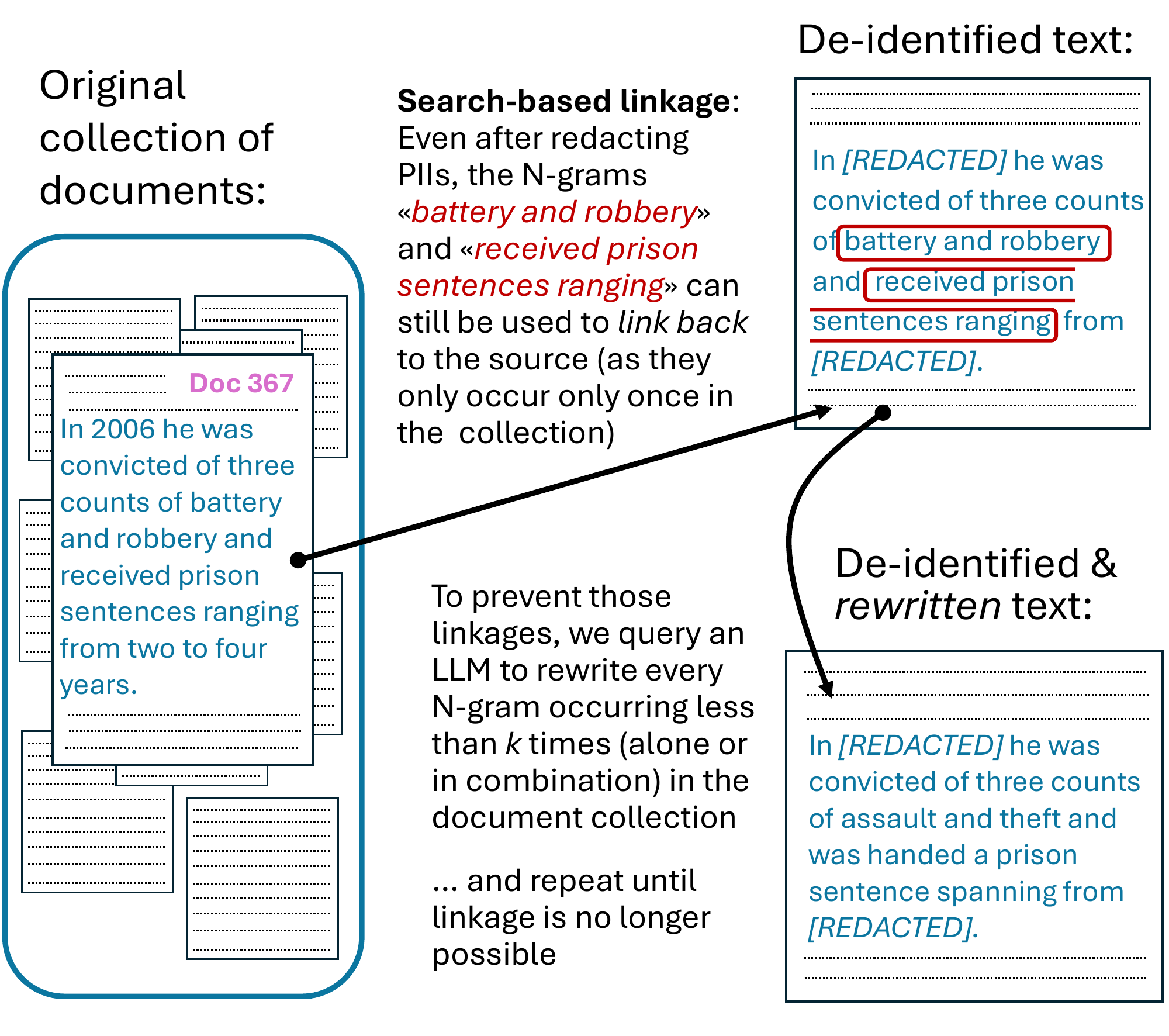}
        \vspace{-5mm}
        \caption{To prevent search-based linkages between a de-identified
        document and its original source, we first identify all N-grams occurring
        less than $k$ times in the collection, and then use an instruction-tuned
        LLM to reformulate each of those N-grams (in their local context) until
        the linkages are averted.}
        % \vspace{-1mm}
        \label{fig:sketch}
    \end{figure}

    De-identification techniques primarily focus on reducing the \textit{re-identification
    risk} -- that is, the risk that an adversary might infer the identity of the
    human individual(s) referred to in the document based on the
    PIIs occurring in the text. However, according to data protection
    regulations such as the European GDPR, data owners should also mitigate the risk
    of \textit{record linkages}, which happens when the de-identified data remains
    linkable to other sources, and in particular to the original data
    collection. 
    
    In Europe, this risk is notably emphasised in the Opinion of the
    Article 29 WP on Anonymisation Techniques \cite{WP29_Opinion_05_2014}, which
    plays a key role in defining how anonymisation and pseudonymisation should be
    interpreted under the GDPR. The Opinion highlights the \textit{(un)linkability}
    of pseudonymised data with its original source as one of the three criteria
    (along with the prevention of singling out and inference) that anonymisation
    techniques should meet. Linkability can indeed allow sensitive information to be traced back to the original records, with potentially harmful consequences for the individuals in question, for instance if de-identified records labelled with health concerns are used by insurance firms to deny medical cover.

    One simple but powerful technique for establishing such linkages is \textit{phrase
    search} \cite{weitzenboeck2022gdpr}. This method entails extracting phrases from
    the de-identified text and then searching for their occurrence in the original
    document collection. As many longer N-grams will only occur once in a given
    corpus, \cite{Silva2020-sv}, phrase search can efficiently connect the
    edited texts back to their source, as illustrated in Fig. \ref{fig:sketch}.

    Phrase search attacks can be conducted by adversaries with knowledge of only
    a few N-grams occurring in the original document. For instance, a de-identified
    patient record $d'$ from a particular hospital may include the mention of a
    rare diagnosis named $X$, or include a particular sentence whose writing style reveals the identity of the doctor in charge of the patient. An adversary with the knowledge that an individual
    diagnosed with $X$ was treated in that hospital (and presumably had $X$
    mentioned in their original patient record $d$) could then infer a link between
    this de-identified record $d'$ and the original patient record $d$.

    This paper presents a method for preventing such search-based linkages through
    an \textit{inverted index} that captures the occurrence of N-grams across
    the collection. This inverted index is employed to efficiently derive the N-grams
    -- and combinations of N-grams -- that are associated with fewer than $k$ documents
    and should therefore be rephrased when they occur in a given document. An
    instruction-tuned LLM is then queried with the text along with this list of
    N-grams to edit. This process is repeated until the text no longer contains any
    N-gram (or combination of N-grams) that could uniquely identify it within the
    original corpus.

    Using two datasets (respectively based on international court cases and
    Wikipedia biographies), we show that existing text rewriting methods are
    poor at preventing search-based linkages. In contrast, the method proposed
    in this paper is able to drastically reduce linkage risks while remaining
    truthful to the original content of the document. However, more
    sophisticated linkage attacks remain possible for adversaries with access to
    the full collection of original documents.

    The paper is structured as follows. Section \ref{sec:related_work} reviews related
    work on text de-identification, record linkages and paraphrasing. Section \ref{sec:approach}
    describes the threat model along with the text rewriting method. This method
    is evaluated in Section \ref{sec:evaluation} on two datasets. Finally,
    Section \ref{sec:discussion} discusses the results, followed by a conclusion
    in Section \ref{sec:conclusion}.

    \section{Related Work}
    \label{sec:related_work}

    \paragraph{Text de-identification}
    The goal of text de-identification, sometimes also called \textit{PII
    removal}, \textit{text sanitisation} or \textit{pseudonymisation}, is to detect
    spans expressing Personally Identifiable Information (PII) and subsequently mask
    those from the documents. Those PIIs may include both direct identifiers, such
    as full person names, phone numbers or home addresses, but also so-called \textit{quasi}-identifiers,
    which do not disclose the person identity in isolation, but may do so when
    combined with other quasi-identifiers and background knowledge. Such quasi-identifiers
    encompass a wide range of information, such as year of birth, professional occupation,
    gender, hometown or nationality. Since both direct identifiers and quasi-identifiers
    occurring in a text may increase the risk of revealing the identities of the
    individuals mentioned, they should generally be masked or redacted.

    Methods for text de-identification may rely on either heuristics or rule-based
    approaches \cite{Neamatullah2008-pu}, information-theoretic measures
    \cite{Sanchez2016-so}, statistical or neural sequence labelling models \cite{Dernoncourt2017-tk,Eder2019-il,pilan-etal-2022-text,Kleinberg2022-lw,Hahm2025-ec},
    unsupervised methods \cite{Morris2022-oo} and more recently LLMs, either simply
    prompted or fined-tuned to the task \cite{Liu2023-tm,Staab2024-ju,Pissarra2024-bh,Frikha2025-pi}. Recent work have also explored how to evaluate the robustness of those methods by attempting the reverse process of re-identification \cite{Patsakis2023-cp,Nyffenegger2024-fs,Charpentier2025-wa,zhang-zhang-2026-de}. Such adversarial evaluations are, however, difficult to conduct on data that is genuinely private, which complicate the assessment of real-world re-identification risks \cite{Ochs2026-zb}.

    Several studies have also proposed privacy-preserving text rewriting methods,
    notably to obfuscate demographic indicators \cite{Xu2019-om} or conceal author
    identity \cite{Weggenmann2022-ro,Bao2024-ku}. Many of those text rewriting methods rely on differential
    privacy (DP) mechanisms \cite{Igamberdiev2023-qs,Utpala2023-hn,Meisenbacher2024-ib}, although alternative approaches have been developed, based on e.g.~adversarial training \cite{Xu2019-om}, tree search \cite{huang-etal-2025-zero} or reinforcement learning \cite{Shi2025-yw}. However,
    those text rewriting techniques produce fully transformed texts that do not
    retain the document's original content. While such transformations are suitable
    for tasks such as synthetic data generation, they are inadequate for many
    downstream uses of text de-identification -- especially in legal or medical contexts
    where faithfulness to the source material is essential. To the best of our
    knowledge, this paper is the first approach specifically designed to prevent
    phrase-based linkage attacks while preserving the document's semantic integrity.

    \paragraph{Record Linkage}

    Record linkage has traditionally been studied in the context of structured datasets,
    where matching relies on deterministic or probabilistic comparisons of
    attributes such as names, birthdates, or postal codes
    \cite{Christen2012-vr,kum-linkage,Vatsalan2013-rm,Gkoulalas-Divanis2021-us}.
    Classical linkage frameworks, such as the Fellegi–Sunter model
    \cite{Fellegi1969-wh}, assume tabular schemas and relatively small sets of attributes,
    making them ill‑suited for unstructured text. Nevertheless, one key idea
    behind those frameworks -- namely that combinations of quasi-identifiers can
    be as revealing as unique identifiers -- carries over to text, in which infrequent
    phrases may act as implicit quasi-identifiers. While a few studies have
    examined record linkage in unstructured data, notably in the medical domain \cite{churi2019systematic},
    the problem remains underexplored. The legal implications of linkage risks in
    unstructured data are analysed in \citet{weitzenboeck2022gdpr}, who highlight
    the difficulty of complying with the GDPR's requirement of unlinkability for
    text or images.

    \paragraph{Paraphrasing}
    Paraphrasing systems have a long history in NLP, with recent progress driven
    by neural approaches and LLMs. These systems are typically fully automated and
    unsupervised \cite{mallinson-etal-2017-paraphrasing,li-etal-2018-paraphrase,witteveen-andrews-2019-paraphrasing,niu-etal-2021-unsupervised,wada-etal-2023-unsupervised}.
    A common limitation, however, is the lack of control over which specific
    phrases are modified and which remain unaltered. As such, standard
    paraphrasing models are poorly equipped to mitigate search-based linkages. Work
    on controllable paraphrasing \cite{Kumar2020-qx,wada-etal-2023-unsupervised}
    introduces some steerability, but are not designed to explicitly replace
    specific substrings for privacy protection, as is the case in the paper.

    \section{Approach}
    \label{sec:approach}

    We first outline the threat scenario, followed by the algorithm used to identify
    the N-grams to rephrase and the LLM-driven rewriting procedure.
    %Appendix \ref{sec:example} provides a step-by-step example.

    \subsection{Threat model}
\label{sec:threat_model}

    We wish to counter privacy threats from adversaries with knowledge of at
    least some segments of the original documents. However, we do not know the
    number and types of substrings that may be available to such adversaries. To
    account for worst-case scenarios, we therefore adopt a maximum-knowledge
    intruder model \cite{domingo2015disclosure} in which the intruder has access
    to the original collection of documents $\{ d \in \mathcal{D}\}$. 
    
    Formally,
    given a de-identified text $d'$, we assume the intruder seeks to establish a
    linkage between $d'$ and the original text $d$ in three consecutive steps:
    \begin{itemize}

        \item Extract one or more substrings $S_{d'}= \{s \mid s \text{ is a contiguous
            sequence of tokens in }d'\}$

        \item Search for the occurrence of those strings in $\mathcal{D}$, which
            results in the subset
            $\mathcal{D}(S_{d'})) = \{d \in \mathcal{D}\mid \forall s \in S_{d'},
            \ s \subseteq d\}$
            of documents containing those strings.

        \item Finally, we consider the linkage as successful if
            $|\mathcal{D}(S_{d'})| < k$, where $k$ is a minimum threshold, as
            done in $k$-anonymity \cite{Samarati1998-zd}. We use both $k=2$ and $k
            =5$ in our experiments.
    \end{itemize}

    Such search-based linkages can be carried out by adversaries with minimal technical
    skills. As discussed in Section \ref{sec:discussion}, more sophisticated types
    of linkages are possible, but mitigating them without incurring large
    distortions to the text content -- and thus undermining the very
    aim of de-identification -- is unfeasible in the general case.

    \subsection{Extraction of N-grams to rephrase}
    \label{sec:ngram_extraction}

    To efficiently determine the frequency of various N-grams (and combinations
    of N-grams) in a document collection $\mathcal{D}$, we rely on an \textit{inverted
    index}, a well-known data structure employed in Information Retrieval. An inverted index maps each word or N-gram to a list (called a \textit{posting
    list}) of documents it appears in. As posting lists are sorted by their
    document identifier, they can be efficiently intersected to derive the list of
    documents where a given combination of N-grams occurs \cite{Manning2008-ir}.

    Once constructed, this index can be used to determine which N-grams
    occurring in a de-identified document $d'$ are associated with fewer than
    $k$ documents in the original collection $\mathcal{D}$, and thus in need of rephrasing
    to prevent search-based linkages.

    \subsubsection{Single N-grams}
    \label{sec:single_ngrams}

    To find a list of single N-grams to rephrase, the following procedure is applied:
    
    \begin{enumerate}

        \item We extract a set $S_{d'}$ of N-grams occurring in $d'$. To reduce
            the number of N-grams to consider, we limit the maximum N-gram length
            (fixed to 8 words in our experiments\footnote{As the frequency of N-grams
            decreases sharply as a function of the N-gram length
            \cite{Silva2020-sv}, considering longer N-grams is usually redundant,
            as those would often encompass shorter N-grams that are themselves
            rare.}), and do not consider N-grams crossing punctuation markers or
            \redacted{} PIIs, which are per definition absent from the original text
            $d$.

        \item We then sort the N-grams in $S_{d'}$ by their length and filter out
            overlapping N-grams, resulting in a set
            $S^{\text{min}}_{d'}\subseteq S_{d'}$ representing the minimal set
            of N-grams that need to be considered as potential linkage points\footnote{We
            select the \textit{shortest} non-overlapping N-grams in $S_{d'}$ to minimise
            the amount of editing needed to prevent linkages. }.

        \item For each N-gram $s \in S^{\text{min}}_{d'}$, we finally check the
            number of documents associated with it in the inverted index
            $\mathcal{I}$. If the length of the posting list $\mathcal{I}(s)$ is
            $< k$, we add $s$ to the N-grams to rephrase.
    \end{enumerate}

    The result of this process is a set $S^{*}_{d'}\subseteq S^{\text{min}}_{d'}$
    of N-grams from $d'$ that are known to be occurring $< k$ times in the original
    corpus of documents and which therefore need to be rephrased.

    \subsubsection{N-gram combinations}
    \label{sec:ngram_combinations}

    To extend the procedure to N-gram \textit{combinations} instead of single N-grams,
    we replace step 3. above with the following two steps:
    \begin{enumerate}

        \item[3b.] We generate the set of possible N-gram combinations $C^{m}_{S^{\text{min}}_{d'}}$
            from the set $S^{\text{min}}_{d'}$, where $m$ is the arity of each
            combination. The maximum arity of the N-gram combinations was set to
            3 in our experiments.

        \item[4b.] For each combination $c \in C^{m}_{S^{\text{min}}_{d'}}$, we
            extract the posting lists of its constituting N-grams and intersect
            them. As the posting lists are already sorted, this intersection can
            be computed in linear time. If the result of this intersection has a
            length $< k$, we add the shortest N-gram in $c$ to the N-grams to
            rephrase.
    \end{enumerate}

    \subsection{In-context Rewriting}

    Once the N-grams $S^{*}_{d'}$ to rephrase in a document $d'$ are extracted,
    we query an instruction-tuned LLM to generate a reformulated text in which
    those N-grams no longer appears \textit{verbatim}:
    \begin{equation}
        d'_{t+1}\leftarrow \mathsf{LLM}(d'_{t}, S^{*}_{d'}) \label{eq:rewriting}
    \end{equation}

    The LLM is provided with the local context surrounding the N-grams
    to rephrase, which helps maintain fluency and contextual integration. To
    scale the rewriting to longer documents and ensure each LLM query is limited
    to a few N-grams, we chunk the text in pieces of a few sentences
    each. The LLM is provided with few-shot examples of rewritten sentences. We also
    found that instructing the LLM to generate a short chain-of-thought reasoning
    ahead of the rewriting led to better results. If rephrasing proves difficult -- for
    instance, if the N-gram contains a named entity --, the LLM is instructed to replace
    it with a placeholder. The full prompt is provided in Appendix
    \ref{sec:prompt}.

    As mentioned in Section \ref{sec:ngram_extraction}, the extraction of N-grams
    is constrained to non-overlapping N-grams to limit the number of N-grams to consider
    in a single rewriting iteration. As a consequence, the rewritten text $d'_{t+1}$
    might still contain some N-grams to rephrase, as those might have been
    ignored due to overlaps. To this end, and to account for the fact that the
    LLM does not always strictly follow its instructions, the rewriting procedure
    is repeated for several passes until the set of N-grams $S^{*}_{d'}$ becomes
    empty.

    \section{Experiments}
    \label{sec:evaluation}

    The approach is empirically evaluated on two datasets and along three
    criteria:
    \begin{itemize}

        \item \textit{Linkage risk}: How many N-grams from the de-identified text
            remain linkable to the original document after rewriting?

        \item \textit{Semantic integrity}: To what extent is the semantic content
            of the original document preserved in the rewritten text?

        \item \textit{Fluency}: To what extent is the rewritten text fluent and natural?
    \end{itemize}

    \subsection{Datasets}

    We rely on two datasets for evaluation, one based on court cases and the
    other on Wikipedia biographies. For each dataset, we start with a large
    collection of original documents $\mathcal{D}$, each document including various
    types of personally identifiable information (PII). We then draw a small
    subset of this collection and use a de-identification tool to produce a
    corresponding set of documents $\mathcal{D}'$ in which PIIs are redacted. This
    de-identification is practically implemented with the Presidio tool\footnote{See
    \href{https://microsoft.github.io/presidio/}{https://microsoft.github.io/presidio/}}.
    As the focus of this paper is on linkage attacks rather than PII detection,
    we do not evaluate here the quality of the de-identification.

    \paragraph{ECHR Court cases}

    We use as first dataset $\mathcal{D}$ a set of 13 759 English-language court
    cases from the
    \href{https://hudoc.echr.coe.int/}{European Court of Human Rights} (ECHR), covering
    the years 1961 to 2020. Court cases constitute an useful testbed for de-identification
    methods, due to their high density in PIIs on various individuals (offenders,
    victims, witnesses, etc.) and the fact that they must carefully balance
    privacy with the need to retain the semantic content of the case. ECHR cases
    have notably previously been employed in the construction of the Text Anonymisation Benchmark (TAB) \cite{pilan-etal-2022-text}.
    A subset $\mathcal{D}'$ of 252 court cases\footnote{This subset corresponds
    to the development and test set of the TAB corpus.} are then drawn from $\mathcal{D}$
    and de-identified with Presidio.

    \paragraph{Wikipedia biographies}

    The second dataset consists of 100 000 biographies sampled from the English Wikipedia.
    The biographies were limited to medium-size lengths (between 800 and 4500 characters),
    and naturally contain a high density of PIIs, such as names, dates, places
    and organisations. As with the court cases, a subset $\mathcal{D}'$ of 250
    biographies are drawn and de-identified with Presidio.

    \subsection{Experimental setup}

\begin{figure}[t!]
            \includegraphics[scale=0.39]{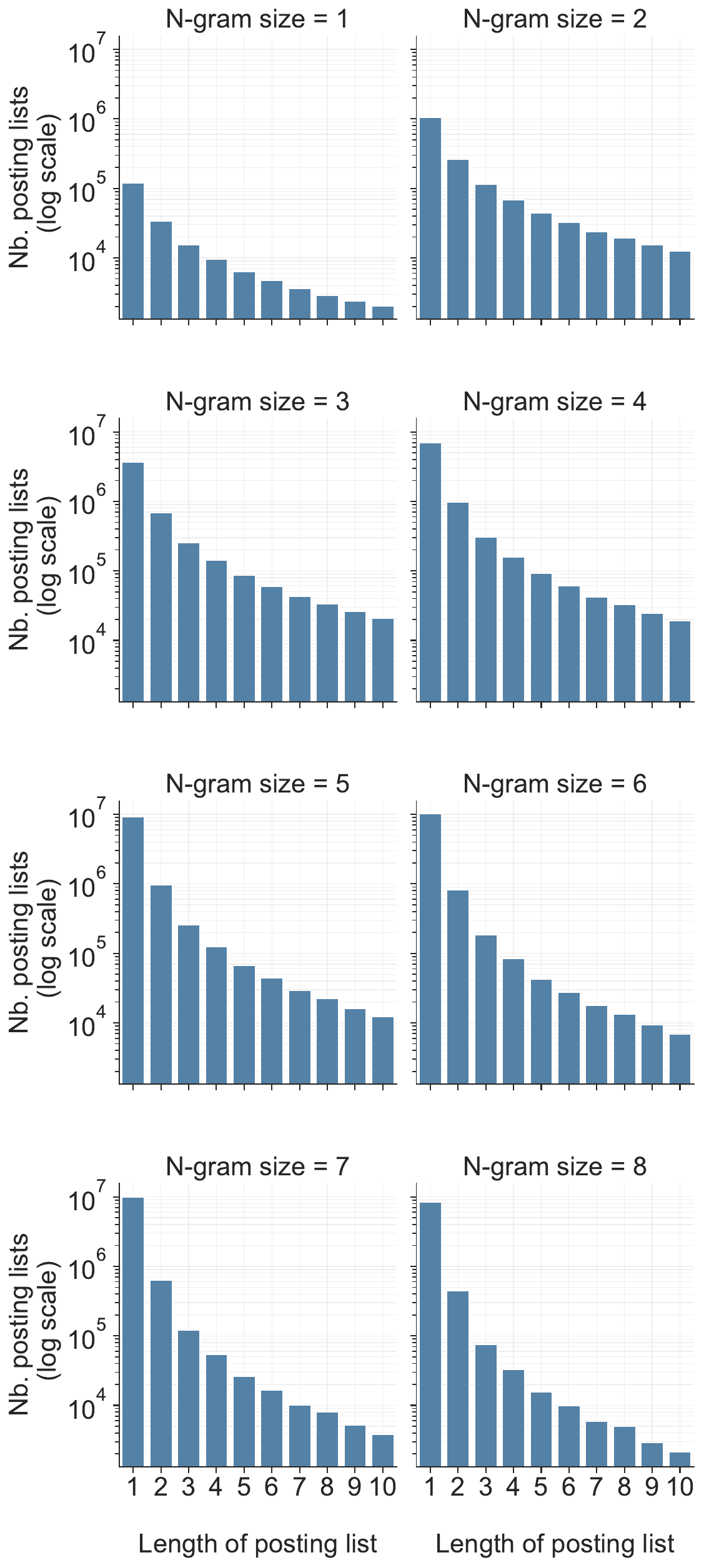}
        
        \caption{Distinct posting lists in the index built from the ECHR court
        cases, distributed according to the posting list length (=nb of documents
        associated with a particular N-gram) and the size of the N-gram, from 1 to
        8 words. The Y-axis is displayed in log scale. Longer N-grams dominate the
        posting lists, as expected since the number of distinct N‑grams grows
        combinatorially with N. The posting lists are in this case quite short,
        particularly for longer N-grams: 60\% of posting lists for unigrams have
        one single document, while this proportion reaches 93\% for $N=8$.}
        \label{fig:posting_list_stats}
        \vspace{-1mm}
    \end{figure}

    As described in Section \ref{sec:approach}, the approach proceeds in several
    steps. An inverted index is initially constructed from the full collection of
    documents $\mathcal{D}$. For the ECHR dataset, the inverted index contains 58.3M
    distinct N-grams (up to 8 words), with an average posting list length of 1.27
    documents ($\pm 0.93$). Figure \ref{fig:posting_list_stats} illustrates the
    distribution of those posting lists in the inverted index.

    Based on this index and a de-identified document $d'$, we derive the
    list of N-grams to rephrase as described in Section
    \ref{sec:ngram_extraction}. Once this list is obtained, we query an
    instruction-tuned LLM to rewrite the text. We used in our experiments a 
    \begin{small}\textsf{\href{https://huggingface.co/Qwen/Qwen3-4B-Instruct-2507}{Qwen3-4B}}
    \end{small}
    model \cite{qwen3technicalreport}, without fine-tuning but with few-shot
    examples to take advantage of the model's in-context learning abilities. The
    full prompts are provided in Appendix \ref{sec:prompt}. The responses
    are generated with a temperature = $1.2$.

    \subsection{Baselines}

    We compare this approach to three alternative text rewriting methods:
    \begin{description}

        \item[Baseline LLM rewriting] An LLM-based rewriter employing the exact same
            model (Qwen3-4B) and prompt as the above approach, but without being
            provided with a specific list of text spans to edit. In other words,
            the LLM is simply instructed to rewrite the full text while
            preserving its meaning. 
            
            The rewriting process is repeated over multiple
            passes to ensure a sufficient number of edits compared to the original
            text. This baseline is used to assess the extent to which standard
            LLM-based paraphrasing can prevent search-based linkages.

        \item[Handcrafted paraphrasing] A rule-based paraphrasing approach that replaces
            nouns and verbs by synonyms provided by Wordnet \cite{miller1995wordnet},
            along with rules for altering modifiers such as adjectives and
            adverbs (see Appendix \ref{sec:controlled_paraphrasing} for details).

        \item[DP-based rewriting] The text rewriting model \cite{Meisenbacher2024-ib},
            which relies on differential privacy (DP) and masked language models
            (MLMs) to obfuscate the text while seeking to preserve textual
            similarity. We use the implementation from the authors, along with
            two $\epsilon$ values 10 and 100, which express the model's \textit{privacy
            budget}.
    \end{description}

    \begin{table*}
        [t!]
        \begin{tabular}{p{60mm}|>{\raggedleft\arraybackslash}p{14mm}>{\raggedleft\arraybackslash}p{14mm}|>{\raggedleft\arraybackslash}p{15mm}>{\raggedleft\arraybackslash}p{12mm}|>{\raggedleft\arraybackslash}p{20mm}}
                                                                                 & \multicolumn{2}{c|}{\textbf{Proportion of}}    & \multicolumn{2}{c|}{\textbf{Semantic }}  & \multirow[t]{2}{2cm}{\centering\textbf{Perplexity}}           \\
                                                                                 & \multicolumn{2}{c|}{\textbf{linkable N-grams}} & \multicolumn{2}{c|}{\textbf{Similarity}} &                                                               \\
            \textbf{Approaches:}                                                 & arity$=1$                                      & arity $\leq 3$                           & \begin{footnotesize}\textsf{distilroberta}\end{footnotesize} & \begin{footnotesize}\centering \textsf{GTE}\end{footnotesize} \  & ( \begin{footnotesize}\centering\textsf{Gemma3}\end{footnotesize})\ \ \ \  \\
            \noalign{\vskip 4pt}                                                  \hline
            \noalign{\vskip 4pt}                                                  %to rephrase (this approach) & & & & \\
            \rule{0pt}{8pt}Baseline LLM-based rewriting                          & 0.597                                          & 0.695                                    & 0.974                                                        & 0.979                                                            & 12.6\\
            (without specified spans to rephrase)                                &                                                &                                          &                                                              &                                                                   \\[1.5mm]
            Handcrafted paraphrasing                                             & 0.348                                          & 0.615                                    & 0.992                                                        & 0.979                                                            & 20.4\\[1.5mm]
            DP-based rewriting ($\epsilon=10$)                                   & 0.001                                          & 0.001                                    & 0.358                                                        & 0.617                                                            & 2736\\
            \rule{0pt}{8pt}DP-based rewriting ($\epsilon=100$)                   & 0.065                                          & 0.130                                    & 0.769                                                        & 0.873                                                            & 171\\[1.5mm]
            \hdashline \noalign{\vskip 4pt} LLM-based rewriting w/ list of spans & 0.001                                          & 0.432                                    & 0.987                                                        & 0.978                                                            & 14.2\\
            (text spans extracted for arity = 1)                                 &                                                &                                          &                                                              &                                                                  &                                                                            \\[1.5mm]
            LLM-based rewriting w/ list of spans                                 & 0.002                                          & 0.002                                    & 0.959                                                        & 0.948                                                            & 15.0                                                                       \\
            (text spans extracted for arity $\leq$ 3)                            &                                                &                                          &                                                              &                                                                  &                                                                            \\
        \end{tabular}
     %   \vspace{-1mm}
        \caption{Evaluation results on the ECHR court cases according
        to three criteria: \textit{linkage risk} (either for single N-grams
        or N-grams combinations of arity $\leq$ 3), \textit{semantic integrity}
        and \textit{fluency}. The first two columns indicate the proportion of N-grams
        that remain linkable to the original document $d$ after rewriting for a
        value $k=2$, which corresponds to the risk of singling out. Semantic
        integrity is measured through cosine similarities between the document
        embeddings of the text before and after rewriting. The embeddings are respectively obtained with
        \begin{scriptsize}
            \textsf{ paraphrase-distilroberta-base-v2}
        \end{scriptsize}
        and
        \begin{scriptsize}
            \textsf{GTE-ModernBERT}
        \end{scriptsize}. Finally, perplexity is measured with
        \begin{scriptsize}
            \textsf{gemma 3-1B-pt}
        \end{scriptsize}. All LLM-based rewriting approaches are based on Qwen3-4B
        (see prompts in Appendix). Results obtained with a single run.}
        \label{table:results}
    \end{table*}

    \subsection{Metrics}

    The three criteria mentioned above (linkage risk, semantic integrity and perplexity) are measured through the following metrics:

    \begin{description}

        \item[Proportion of linkable N-grams] The likelihood of search-based linkages is computed by
            counting the number of N-grams from the document $d'$ that can univocally
            link back to $d$, and measuring the proportion of those N-grams that
            remain present after rewriting. We assess both linkages with single N-grams
            (arity=1) as well as linkages from N-gram combinations (with an
            arity $\leq 3$).

        \item[Semantic similarity] We measure the extent to which the semantic content
            of  $d'$ is preserved in the rewritten text through cosine similarities
            between their document embeddingss, as extracted through two encoder-only
            language models:
            \begin{small}
                \textsf{ \href{https://huggingface.co/sentence-transformers/paraphrase-distilroberta-base-v2}{paraphrase-distilroberta-base-v2}}
            \end{small}
            and
            \begin{small}
                \textsf{\href{https://huggingface.co/Alibaba-NLP/gte-modernbert-base}{GTE-ModernBERT}}
            \end{small}.

        \item[Perplexity] Finally, perplexity is used to measure the linguistic
            quality of the rewritten document. We use the
            \begin{small}
                \textsf{\href{https://huggingface.co/google/gemma-3-1b-pt}{gemma
                3-1B-pt}}
            \end{small}
            base model \cite{Gemma-Team2025-hu} for this purpose.
    \end{description}

    \begin{table*}
        [t!]
        \begin{tabular}{p{60mm}|>{\raggedleft\arraybackslash}p{14mm}>{\raggedleft\arraybackslash}p{14mm}|>{\raggedleft\arraybackslash}p{15mm}>{\raggedleft\arraybackslash}p{12mm}|>{\raggedleft\arraybackslash}p{20mm}}
                                                                                 & \multicolumn{2}{c|}{\textbf{Proportion of}}    & \multicolumn{2}{c|}{\textbf{Semantic }}  & \multirow[t]{2}{2cm}{\centering\textbf{Perplexity}}           \\
                                                                                 & \multicolumn{2}{c|}{\textbf{linkable N-grams}} & \multicolumn{2}{c|}{\textbf{Similarity}} &                                                               \\
            \textbf{Approaches:}                                                 & arity$=1$                                      & arity $\leq 3$                           & \begin{footnotesize}\textsf{distilroberta}\end{footnotesize} & \begin{footnotesize}\centering \textsf{GTE}\end{footnotesize} \  & ( \begin{footnotesize}\centering\textsf{Gemma3}\end{footnotesize})\ \ \ \  \\
            \noalign{\vskip 4pt}                                                  \hline
            \noalign{\vskip 4pt}                                                  %to rephrase (this approach) & & & & \\
            \rule{0pt}{8pt}Baseline LLM-based rewriting                          & 0.485                                          & 0.553                                    & 0.989                                                        & 0.994                                                           & 10.7\\
            (without specified spans to rephrase)                                &                                                &                                          &                                                              &                                                                   \\[1.5mm]
            Handcrafted paraphrasing                                             & 0.256                                          & 0.355                                    & 0.952                                                        & 0.978                                                            & 19.17\\[1.5mm]
            DP-based rewriting ($\epsilon=10$)                                   & 0.008                                          & 0.009                                    & 0.268                                                        & 0.608                                                            & 2122\\
            \rule{0pt}{8pt}DP-based rewriting ($\epsilon=100$)                   & 0.051                                          & 0.091                                    & 0.618                                                        & 0.808                                                            & 136                                                                        \\[1.5mm]
            \hdashline \noalign{\vskip 4pt} LLM-based rewriting w/ list of spans &                     0.010                           &           0.346                               &       0.841                                                       &         0.914                                                         & 12.9\\
            (text spans extracted for arity = 1)                                 &                                                &                                          &                                                              &                                                                  &                                                                            \\[1.5mm]
            LLM-based rewriting w/ list of spans                                 & 0.002                                          & 0.003                                    & 0.733                                                        & 0.858                                                            & 14.7\\
            (text spans extracted for arity $\leq$ 3)                            &                                                &                                          &                                                              &                                                                  &                                                                            \\
        \end{tabular}
    %    \vspace{-1mm}
        \caption{Evaluation results on the dataset of Wikipedia biographies
        according to same metrics as Table \ref{table:results}.}
        \label{table:results2}
    \end{table*}

    \subsection{Results}
    Tables \ref{table:results} and \ref{table:results2} present the results for the
    two datasets. As we can observe, the baseline LLM rewriting -- which simply
    rewrites the full text without being provided with specific text spans to
    edit -- preserves most of the semantic content but only hinders 40-50\% of
    direct N-gram linkages. Handcrafted paraphrasing leads to modest
    numbers of edits, but with only 65-75\% of direct linkages removed, and an increase in perplexity due to suboptimal lexical choices.
    Finally, DP-based rewriting perform poorly, producing disfluent texts
    with weak semantic ties to the source. In contrast, the proposed method (last
    two lines in the Tables) eliminates around 99.8\% of N-gram
    linkages, although a small number persists when the LLM is repeatedly unable to
    find suitable rephrasing.

    We can observe a lower semantic similarity and higher perplexity of the rewritten
    biographies compared to the court cases. This seems to be due to the fact
    that many of the spans to replace in the biographies correspond to named
    entities or other highly specific terms that have a high impact on the
    document semantic similarity.

    The two Tables focus on linkages with $k=2$, which corresponds to the risk
    of singling out. The approach can be applied to higher values of $k$, at the
    cost of more extensive edits, leading itself to slight drops in semantic
    similarity\footnote{For instance, with $k=5$ instead of $k=2$, the semantic similarity of the
    presented approach (as measured by \begin{scriptsize}\textsf{distilroberta}\end{scriptsize} on the ECHR court
    cases) drops from 0.987 to 0.975.}.

    Appendix \ref{sec:example} provides a full, step-by-step example of a court case document rewritten
    using the baselines and proposed approach.

    In terms of computing time, the method took (on a single GPU) an average of
    15.5 ($\pm 12.8$) secs per document to avert linkages with single N-grams, and
    52.6 ($\pm 37.2$) secs to avert linkages with N-gram combinations (up to
    arity 3). The court cases have an average of 750 ($\pm 389$) words.

    \section{Discussion}
    \label{sec:discussion}

    Overall, the proposed approach seems to provide the most effective balance
    between precluding search-based linkages and preserving the semantic content
    of the document. The preserved content is, however, influenced by the nature
    of the texts: while the edits on the court cases were mostly focused on
    general phrases that could be easily reformulated, many edits on the Wikipedia
    biographies revolved around named entities and other specific, harder-to-rephrase
    terms, thereby leading to a larger drop in semantic similarity.

    While the approach limited to the rephrasing of
    single "linkable" N-grams (arity=1) produces a rewritten text that is
    semantically closer to its input, it fails against linkages due to N-gram combinations.
    As shown in the results, those indirect linkages can also be removed (arity$\leq
    3$), although at the cost of more extensive edits, leading to a slight drop
    in semantic similarity.

    One important limitation of the presented approach is its focus on linkage
    attacks implemented through \textit{phrase search}. This focus on phrase
    search linkages is motivated by two considerations:
    \begin{itemize}

        \item Search-based linkage attacks can be carried out by adversaries with
            minimal technical skills along with the knowledge of some text segments (N-grams)
            of the original texts.

        \item Those linkages can also be averted while retaining the core semantic content of the text.
    \end{itemize}

    However, more sophisticated attacks are possible. As evidenced in the
    results, LLM-based rewriting produce texts that remain semantically close to the initial document. While this characteristic is desirable from a data
    utility perspective, it also suggest that linkage attacks can be conducted at
    the level of \textit{latent representations} (such as embedding vectors) instead of surface forms.

    Indeed, we observed in our two datasets that when computing cosine similarities
    between $d'$ and the original document collection $\mathcal{D}$, the original
    document $d$ ranked first in the vast majority of cases, and that this
    proximity remained fairly stable across rewriting iterations, even after many
    rewriting passes. The proposed method does not, as a consequence, provide a
    robust protection against such semantics-based linkages.

    Could semantics-based linkage attacks be prevented through further edits? In theory, yes, but mitigating those attacks would, by definition,
    require altering the semantic content of the document to such an extent that
    it could no longer be related to the original text. Such large distortions
    would undermine the very aim of text de-identification methods, and are therefore
    difficult to justify from a data utility perspective.

\begin{figure}[t]
        \begin{center}
        \includegraphics[scale=0.62]{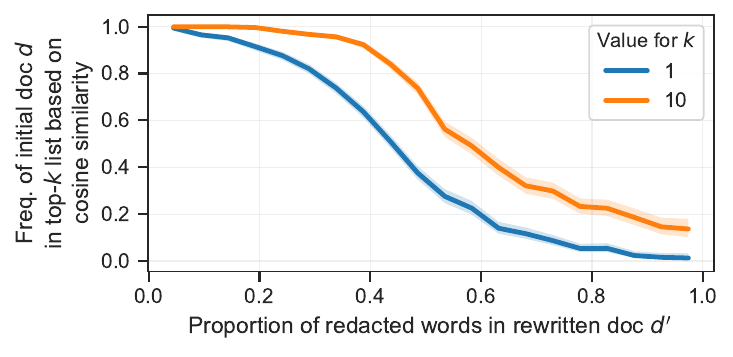}
        \end{center} \vspace{-4mm}
        \caption{Frequency of original text $d$ appearing in the top-$k$ list among all documents in
        collection $\mathcal{D}$ when sorted on cosine similarity with rewritten
        doc $d'$, as a function of the proportion of words redacted in $d'$. The
        results are averaged on the dataset of court cases. The shaded
        areas represent the Wilson CI. We can observe 
        that the original text $d$ remains semantically very close to the
        rewritten document $d'$ until a large proportion of words (around 50\%) are
        redacted.}
        \label{fig:linkage_by_redact_prop}
        \vspace{-2mm}
    \end{figure}

    This difficulty can be verified empirically through a simple experiment where
    we take a text $d$ and gradually redact a small proportion of its words,
    step by step, until all words are edited out. At each iteration, we measure
    the cosine similarity between the initial collection $\mathcal{D}$ and the redacted text $d'$. We can then measure the proportion of words
    that need to be redacted in order to prevent semantics-based linkages --
    that is, to ensure the cosine similarity between $d$ and $d'$ is no longer among
    the top-$k$ highest similarities for $\mathcal{D}$. The results for the ECHR
    court cases, shown in Figure \ref{fig:linkage_by_redact_prop}, demonstrate
    that the prevention of semantics-based linkages requires drastic alterations
    to the document, with at least half the words needing to be redacted\footnote{Alternatively,
    one could also increase the distance between $d$ and $d'$ by adding,
    removing or substituting segments of the text. This strategy would, however,
    conflict with the need to remain faithful to the original content of the text,
    which lies at the core of de-identification methods.}. Needless to say, such
    level of obfuscation is hardly compatible with most de-identification use
    cases, as the resulting texts are barely readable.

    However, semantics-based linkages are arguably a less pressing concern than search-based
    attacks. Although the threat model of Section \ref{sec:threat_model} assumed a worst-case scenario with a maximum-knowledge intruder, search-based linkages can in practice be conducted
    based on the knowledge of a limited set of N-grams from the original texts. 
    Semantics-based linkages, on the other hand, is conditioned on adversaries with access to the original documents in full text in order to compute the corresponding embedding vectors -- an assumption that is unlikely to hold in practical adversarial scenarios.

    \section{Conclusion}
    \label{sec:conclusion}

    Text de-identification methods primarily focus on the detection and redaction
    of personally identifiable information (PII), with the objective of reducing
    the risk of re-identifying the individuals mentioned or referred to in the
    text. However, data protection regulations such as the GDPR also emphasise
    the need to mitigate the risk of \textit{record linkages}, which can occur
    when an adversary is able to link a de-identified document to its original
    counterpart in the source database.

    A particularly simple method of record linkage is through \textit{phrase
    search}, where an adversary links the de-identified document to its original
    counterpart based on the occurrence of specific phrases (i.e., N-grams) that
    are present in the original text and remain unchanged in the de-identified
    version. We proposed a method to prevent such search-based linkages based on the construction
    of an inverted index, making it possible to determine the
    documents in which any particular N-gram appears. Given de-identified document $d'$, the method extracts the N-grams appearing
    in fewer than $k$ documents in the original collection and prompts an LLM to
    rephrase the text such that those N-grams no longer appear. The process is repeated
     until all problematic N-grams are rewritten.

     Beyond preventing phrase linkages, the rewriting operations also acts as a broader \textit{privacy-enhancing measure}, as they effectively limit the occurrence of rare or atypical terms in de-identified documents. Due to their rarity (relative to their underlying document collection), such distinctive terms may indeed be exploited by adversaries to infer additional cues about the potential identity of the individual(s), even when those terms do not correspond to PIIs in the narrow sense.
     
    Experiments on two datasets demonstrate that the method can successfully
    mitigate the risk of search-based linkages while maintaining the core content
    and fluency of the rewritten texts. However, as mentioned above,
    more sophisticated attacks remain possible, notably based on latent semantic
    representations derived from the texts.

    \section*{Limitations}
    \label{sec:limitations}

    As covered in the discussion, the proposed text rewriting approach is focused
    on preventing search-based linkages rather than linkages based at the level of
    latent representations such as document embeddings. As a consequence, the method
    does not prevent the possibility of semantic linkages.

    Another assumption that underlies the proposed approach is that we have access
    to a collection $\mathcal{D}$ of original documents that constitute a "pool"
    from which the de-identified documents are drawn and subsequently edited. However,
    it should be noted that, if no such collection is available, the risk of
    linkages disappears, and hence the need to protect de-identified documents against
    those.

    Due to space constraints, the evaluation of the approach is limited to two
    datasets, written in a single language (English). While we expect that those
    results also carry over other types of documents and languages beyond those
    datasets, this remains to be tested experimentally. The same comment applies
    to the use of other model architectures and parameter size than the one
    employed for the evaluation (Qwen3-4B).

    \section*{Ethical considerations}

    While the proposed method aims to strengthen the privacy of de-identified texts
    by reducing their traceability to the original database, there is an
    inherent risk that such approach might give a false sense of security of the
    level of privacy it provides. Indeed, although de-identification can help
    conceal a large number of PIIs expressed in a document, they remain imperfect
    and prone to various errors and omissions \cite{Sarkar2024-ay}, even when
    combined with the protection against search-based linkage attacks presented
    in this paper. It should therefore be stressed that de-identification is a privacy-enhancing
    technique but does not provide a strong privacy guarantee or equate full anonymisation.

\section*{Acknowledgements}

This publication has been funded by the SFI NorwAI, (Centre for Research-based Innovation, 309834). The authors gratefully acknowledge the financial support from the Research Council of Norway.

    \bibliography{biblio}

    \appendix

\newpage

\onecolumn

\section{Appendix}
\label{sec:appendix}

\subsection{Prompt}
\label{sec:prompt}

The LLM prompt employed to rephrase N-grams in a given input text (see Eq. \ref{eq:rewriting})
is provided below.

\begin{tcolorbox}
    [enhanced, width=1.0\textwidth, colback=gray!5, colframe=gray!80, boxrule=0.8pt,
    breakable, title=LLM rewriting prompt, before skip=10pt, after skip=10pt]

    As a text rewriting tool, you are given a text along with a list of spans occurring
    in that text. You must edit the text such that EVERY span in the list is
    replaced. The replacements must ensure that the original span no longer appears
    as such in the text. But the changes can be small, such as removing or inserting
    an adjective or adverb, using a close synonym, paraphrasing part of the span,
    or changing verb tense. The overall meaning should be retained as much as possible.
    You can also rewrite the words around the span if needed.
    \vspace{3mm}

    Many named entities have already been replaced by \redacted{}, but not all. If
    one of the spans to replace is a named entity (person name, place,
    organisation, acronym, etc.) or another highly specific term impossible to
    rephrase, you should replace it by \redacted{}. Never attempt to replace an entity
    by inventing a new name.
    \vspace{3mm}

    Only use \redacted{} if there are really no other options to edit the span (if
    the span is larger than the entity, you should first edit the other words),
    and avoid redacting article numbers.
    \vspace{3mm}

    \textbf{Examples}:

    \textit{Text}: The quick brown fox jumps over the lazy dog.

    \textit{Span(s) to replace}: "quick brown fox", "over the lazy dog"

    \textit{Output}: The fast brown fox jumps above the lazy dog.
    \vspace{3mm}

    \textit{Text}: The Kalininskiy District Court of Appeal upheld the decision of
    the lower court.

    \textit{Span(s) to replace}: "Kalininskiy District Court of Appeal", "of the
    lower court"

    \textit{Output}: The \redacted{} Court of Appeal upheld the decision made by
    the lower court.
    \vspace{3mm}

    \textit{Text}: The applicant was charged with theft and sentenced to five
    years in prison.

    \textit{Span(s) to replace}: "theft", "five years in prison"

    \textit{Output}: The applicant was accused of stealing and given a five-year
    prison term.
    \vspace{3mm}

    \textit{Text}: The appeal was filed on June 5, 2020 to the District Court of
    Novosibirsk"

    \textit{Span(s) to replace}: "The appeal", "June 5, 2020", "of Novosibirsk"

    \textit{Output}: This appeal was filed on June 5th, 2020 to the District
    Court of \redacted{}.
    \vspace{3mm}

    \textit{Text}: The judge reviewed the evidence and ruled in favour of the prosecution.

    \textit{Span(s) to replace}: "reviewed the evidence", "ruled in favour of the"

    \textit{Output}: The judge reviewed the available evidence and decided to side
    with the prosecution.
    \vspace{3mm}

    \textit{Text}: The judge examined the witness statements and delivered his final
    judgment.

    \textit{Span(s) to replace}: "examined the witness statements", "delivered his
    final"

    \textit{Output}: The witness statements were examined by the judge, who
    delivered afterwards his final judgment.
    \vspace{3mm}

    You should first output a short reasoning (max 100 words) and then output a
    JSON object with one single field, 'edited\_text', containing the entire
    modified text with ALL the given spans replaced by an alternative.
    \vspace{3mm}

    Now, here is the text to edit and the spans to replace:

    \textit{Text}: \slotbox{TEXT}

    \textit{Span(s) to replace}: \slotbox{SPANS\_LIST}
\end{tcolorbox}

%\newpage
\subsection{Controlled paraphrasing}
\label{sec:controlled_paraphrasing}

One of the 3 baselines employed in the evaluation (Section \ref{sec:evaluation})
relies on a controlled paraphrasing approach. We provide below further technical
details on this baseline.

\paragraph{Edit Functions}
The documents are initially processed using the UDPipe parser
\cite{Straka2018-ct} to extract part-of-speech tags and morphological features. Edits
are restricted to content words—verbs, adverbs, nouns, and adjectives. For each category,
we consult the WordNet lexical database \cite{miller1995wordnet} to identify
synonyms. This constitutes the first edit function, \texttt{replacement}. The
second, \texttt{removal}, applies only to adjectives and adverbs. These
functions are applied to each unique phrase in the development documents of the
TAB corpus, generating a set of candidate replacements for any phrase requiring alteration.
For \texttt{replacement}, the original word’s morphological state guides the
transformation of the WordNet lemma into the correct form. For instance, given
the phrase \textit{he was charged}, the lemma \textit{charge} is searched as a verb,
yielding a synonym such as \textit{accuse}, which is then conjugated to \textit{accused}.
This process ensures correct verb forms, noun number, and adjective/adverb
comparison degrees, using the Pattern Python library \cite{de2012pattern}. 

\paragraph{Replacement Selection}
To preserve semantic integrity while altering document form, we apply a weighted
scoring method. Levenshtein distance quantifies form differences between
original and candidate phrases. Semantic similarity is assessed using cosine similarity
between embeddings generated by a sentence transformer model\footnote{\href{https://huggingface.co/sentence-transformers/bert-base-nli-mean-tokens}{https://huggingface.co/sentence-transformers/bert-base-nli-mean-tokens}.}
\cite{zhang2019bertscore}. The harmonic mean of these scores determines the optimal
replacement.

%\newpage
\subsection{Full example}
\label{sec:example}

\subsubsection{Inputs}

We provide here a full example, starting with an initial ECHR court case\footnote{Source:
\href{https://hudoc.echr.coe.int/eng?i=001-159913}{Prus v. Poland, no. 5136/11,
ECHR 2016.}} from the document collection:

\begin{tcolorbox}
    [enhanced, width=1.0\textwidth, colback=gray!5, colframe=gray!80, boxrule=0.8pt,
    breakable, title=Original court case $d$, before skip=10pt, after skip=10pt]
    %\begin{small}
    \textbf{PROCEDURE}
    \vspace{1mm}

    1. The case originated in an application (no. 5136/11) against the Republic
    of Poland lodged with the Court under Article 34 of the Convention for the
    Protection of Human Rights and Fundamental Freedoms (“the Convention”) by a
    Polish national, Mr Kamil Prus (“the applicant”), on 6 December 2010.
    \vspace{1mm}

    2. The applicant, who had been granted legal aid, was represented by Mr S.
    Kotuła, a lawyer practising in Lublin. The Polish Government (“the Government”)
    were represented by their Agent, Ms J. Chrzanowska, of the Ministry of Foreign
    Affairs.
    \vspace{1mm}

    3. The applicant alleged, in particular, a breach of Article 3 of the Convention
    on account of the imposition on him of the so-called “dangerous detainee” regime.
    \vspace{1mm}

    4. On 7 July 2014 the application was communicated to the Government.
    \vspace{3mm}

    \textbf{THE FACTS }
    \vspace{1mm}

    \textbf{I. THE CIRCUMSTANCES OF THE CASE }
    \vspace{1mm}

    5. The applicant was born in 1987 and is detained in Lublin.
    \vspace{2mm}

    \textbf{A. Criminal proceedings against the applicant}
    \vspace{1mm}

    6. On 1 December 2005 the applicant was arrested and remanded in custody. In
    2006 he was convicted of three counts of battery and robbery and received
    prison sentences ranging from two to four years. On 28 November 2006 the Lublin
    Regional Court imposed a cumulative sentence for four criminal convictions.
    \vspace{2mm}

    \textbf{B. Imposition of the so-called “dangerous detainee” regime}
    \vspace{1mm}

    7. On 5 November 2010 the Opole Lubelskie Prison Penitentiary Commission imposed
    on the applicant the so-called “dangerous detainee” regime. The commission
    held that the applicant had been the leader of or an active participant in a
    planned collective remonstrance in Opole Lubelskie Prison (Article 88a § 2 (2a)
    of the Code of Execution of Criminal Sentences), as the detainees refused to
    accept food provided by the prison authorities. The authorities learned that
    the detainees were planning another collective protest for 11 November 2010.
    The commission also considered that the applicant was particularly depraved.
    \vspace{1mm}

    8. On 18 January 2011 the Opole Lubelskie Prison Penitentiary Commission
    rejected an appeal lodged by the applicant and dismissed his request for
    leave to examine the appeal outside the statutory time-limit.
    \vspace{1mm}

    9. Subsequently, the applicant was transferred to the Lublin Remand Centre.
    \vspace{1mm}

    10. On 3 February 2011 the Lublin Remand Centre Penitentiary Commission reviewed
    and upheld the decision to apply the regime to the applicant, considering
    that he posed a serious threat to the security of the detention centre. The applicant
    appealed.
    \vspace{1mm}

    11. On 4 April 2011 the Lublin Regional Court dismissed the appeal. The
    court referred to the reasons given in the decision of 5 November 2010, holding
    that the decision had been lawful and justified.
    \vspace{1mm}

    12. On 3 March 2011 the applicant lodged a complaint with the Lublin Regional
    Inspectorate of Prisons. He complained that he was having difficulties in
    accessing educational activities in the Lublin Remand Centre. The authorities
    dismissed the complaint, holding that educational activities were not
    organised for dangerous prisoners.
    \vspace{1mm}

    13. On an unspecified date in 2011 the applicant notified the police of his
    allegedly unlawful classification as a “dangerous detainee” on 5 November
    2010. He claimed that the Opole Lubelskie Prison authorities had exceeded
    their powers in imposing the regime on him. On 29 April 2011 the Opole Lubelskie
    District Prosecutor refused to open an investigation in the case, holding
    that no offence had been committed.
    \vspace{1mm}

    14. On 28 April 2011 the Lublin Remand Centre Penitentiary Commission
    revoked its decision to classify the applicant as a “dangerous detainee”. It
    considered that the applicant’s behaviour had improved and that he no longer
    posed a threat to the security of the remand centre.

    %\end{small}
\end{tcolorbox}

%\newpage

Assume this initial court case has been de-identified, resulting in a new
document $d'$. The de-identification was here conducted with the Presidio de-identification
tool\footnote{As one can observe in the document below, some text spans that were
masked by the tool could probably have been kept in clear text (such as `Republic
of Poland') while some actual PIIs were left out of the de-identification. Those
errors and omissions do not, however, affect in any way the operation of the
proposed approach.}.

\begin{tcolorbox}
    [enhanced, width=1.0\textwidth, colback=gray!5, colframe=gray!80, boxrule=0.8pt,
    breakable, title=De-identified court case $d'$ , before skip=10pt, after skip=10pt]
    %\begin{small}
    \textbf{PROCEDURE}
    \vspace{1mm}

    1. The case originated in an application (no. \redactedsmall{}) against
    \redactedsmall{} lodged with the Court under Article 34 of the Convention for
    the Protection of Human Rights and Fundamental Freedoms (“the Convention”) by
    a \redactedsmall{} national, Mr \redactedsmall{} (“the applicant”), on \redactedsmall{}.
    \vspace{1mm}

    2. The applicant, who had been granted legal aid, was represented by
    \redactedsmall{}, a lawyer practising in Lublin. The Polish Government (“the
    Government”) were represented by their Agent, \redactedsmall{}, of the
    Ministry of Foreign Affairs.
    \vspace{1mm}

    3. The applicant alleged, in particular, a breach of Article 3 of the
    Convention on account of the imposition on him of the so-called “dangerous
    detainee” regime.
    \vspace{1mm}

    4. On \redactedsmall{} the application was communicated to the Government.
    \vspace{3mm}

    \textbf{THE FACTS}
    \vspace{1mm}

    \textbf{I. THE CIRCUMSTANCES OF THE CASE}
    \vspace{1mm}

    5. The applicant was born in \redactedsmall{} and is detained in
    \redactedsmall{}.
    \vspace{2mm}

    \textbf{A. Criminal proceedings against the applicant}
    \vspace{1mm}

    6. On \redactedsmall{} the applicant was arrested and remanded in custody. In
    \redactedsmall{} he was convicted of three counts of battery and robbery and
    received prison sentences ranging from \redactedsmall{}. On \redactedsmall{}
    the Lublin Regional Court imposed a cumulative sentence for four criminal convictions.
    \vspace{2mm}

    \textbf{B. Imposition of the so-called “dangerous detainee” regime}
    \vspace{1mm}

    7. On \redactedsmall{} the Opole Lubelskie Prison Penitentiary Commission
    imposed on the applicant the so-called “dangerous detainee” regime. The commission
    held that the applicant had been the leader of or an active participant in a
    planned collective remonstrance in \redactedsmall{} (Article 88a § 2 (2a) of
    the Code of Execution of Criminal Sentences), as the detainees refused to accept
    food provided by the prison authorities. The authorities learned that the
    detainees were planning another collective protest for \redactedsmall{}. The
    commission also considered that the applicant was particularly depraved.
    \vspace{1mm}

    8. On \redactedsmall{} the Opole Lubelskie Prison Penitentiary Commission rejected
    an appeal lodged by the applicant and dismissed his request for leave to examine
    the appeal outside the statutory time-limit.
    \vspace{1mm}

    9. Subsequently, the applicant was transferred to the Lublin Remand Centre.
    \vspace{1mm}

    10. On \redactedsmall{} the Lublin Remand Centre Penitentiary Commission reviewed
    and upheld the decision to apply the regime to the applicant, considering
    that he posed a serious threat to the security of the detention centre. The applicant
    appealed.
    \vspace{1mm}

    11. On \redactedsmall{} the Lublin Regional Court dismissed the appeal. The court
    referred to the reasons given in the decision of \redactedsmall{}, holding that
    the decision had been lawful and justified.
    \vspace{1mm}

    12. On \redactedsmall{} the applicant lodged a complaint with the Lublin
    Regional Inspectorate of Prisons. He complained that he was having difficulties
    in accessing educational activities in the Lublin Remand Centre. The
    authorities dismissed the complaint, holding that educational activities were
    not organised for dangerous prisoners.
    \vspace{1mm}

    13. On an unspecified date in \redactedsmall{} the applicant notified the
    police of his allegedly unlawful classification as a “dangerous detainee” on
    \redactedsmall{}. He claimed that the Opole Lubelskie Prison authorities had
    exceeded their powers in imposing the regime on him. On \redactedsmall{} the
    Opole Lubelskie District Prosecutor refused to open an investigation in the
    case, holding that no offence had been committed.
    \vspace{1mm}

    14. On \redactedsmall{} the Lublin Remand Centre Penitentiary Commission
    revoked its decision to classify the applicant as a “dangerous detainee”. It
    considered that the applicant’s behaviour had improved and that he no longer
    posed a threat to the security of the remand centre.

    %\end{small}
\end{tcolorbox}

\subsubsection{Proposed approach}

Based on the approach presented in Section \ref{sec:ngram_extraction}, we can use
the inverted index to derive the N-grams occurring only once or twice ($k =2$) in
the document collection:

\begin{tcolorbox}
    [enhanced, width=1.0\textwidth, colback=gray!5, colframe=gray!80, boxrule=0.8pt,
    breakable, title=N-grams to rephrase ($< 2$ occurrences in $\mathcal{D}$), before
    skip=10pt, after skip=10pt]
    %\begin{small}

    ['battery and robbery', 'and received prison sentences ranging', 'Lublin
    Regional Court imposed', 'for four criminal', 'Lubelskie Prison Penitentiary
    Commission imposed', 'The commission held that the applicant had', 'leader
    of or', 'active participant in a', 'the detainees refused', 'accept food provided',
    'The authorities learned', 'planning another', 'collective protest for', 'commission
    also considered', 'particularly depraved', 'Commission rejected an', 'by the
    applicant and dismissed his', 'for leave to examine', 'the appeal outside
    the', 'reviewed and upheld the decision', 'apply the regime', 'that he posed
    a serious', 'security of the detention', 'court referred to the reasons given
    in the', 'Inspectorate of Prisons', 'he was having difficulties in', 'accessing
    educational', 'activities in the Lublin', 'The authorities dismissed', 'holding
    that educational', 'not organised for', 'police of his allegedly', 'unlawful
    classification', 'detainee” on', 'claimed that the Opole', 'Lubelskie Prison
    authorities', 'powers in imposing', 'Lubelskie District Prosecutor refused',
    'holding that no offence', 'Commission revoked its', 'It considered that the
    applicant’s behaviour', 'improved and that he no']

    %\end{small}
\end{tcolorbox}

After querying the LLM to rewrite the N-grams above, we obtain the following results:

\begin{tcolorbox}
    [enhanced, width=1.0\textwidth, colback=gray!5, colframe=gray!80, boxrule=0.8pt,
    breakable, title=De-identified court case $d'$ after rewriting N-grams with
    $< 2$ occurrences in $\mathcal{D}$, before skip=10pt, after skip=10pt]
    %\begin{small}
    \textbf{PROCEDURE }
    \vspace{1mm}

    1. The case originated in an application (no. \redactedsmall{}) against \redactedsmall{}
    lodged with the Court under Article 34 of the Convention for the Protection
    of Human Rights and Fundamental Freedoms (“the Convention”) by a
    \redactedsmall{} national, Mr \redactedsmall{} (“the applicant”), on
    \redactedsmall{}.
    \vspace{1mm}

    2. The applicant, who had been granted legal aid, was represented by \redactedsmall{},
    a lawyer practising in Lublin. The Polish Government (“the Government”) were
    represented by their Agent, \redactedsmall{}, of the Ministry of Foreign Affairs.
    \vspace{1mm}

    3. The applicant alleged, in particular, a breach of Article 3 of the Convention
    on account of the imposition on him of the so-called “dangerous detainee” regime.
    \vspace{1mm}

    4. On \redactedsmall{} the application was communicated to the Government.
    \vspace{2mm}

    \textbf{THE FACTS }
    \vspace{1mm}

    \textbf{I. THE CIRCUMSTANCES OF THE CASE }
    \vspace{1mm}

    5. The applicant was born in \redactedsmall{} and is detained in \redactedsmall{}.
    \vspace{2mm}

    \textbf{A. Criminal proceedings against the applicant }
    \vspace{1mm}

    6. On \redactedsmall{} the applicant was arrested and remanded in custody.
    In \redactedsmall{} he was convicted of three counts of assault and theft and
    was handed a prison sentence ranging from \redactedsmall{}. On \redactedsmall{}
    the Lublin Regional Court handed down a cumulative sentence across four
    offenses.
    \vspace{2mm}

    \textbf{B. Imposition of the so-called "dangerous detainee" regime }
    \vspace{1mm}

    7. On \redactedsmall{} the Opole Lubelskie Prison Penitentiary Commission ordered
    on the applicant the so-called "dangerous detainee" regime. The commission
    concluded that the applicant had been the head of or involved in a
    coordinated group action in \redactedsmall{} (Article 88a § 2 (2a) of the
    Code of Execution of Criminal Sentences), as the detainees declined the
    meals supplied by the prison authorities. The authorities discovered that the
    detainees were organizing a new group protest for \redactedsmall{}. The commission
    also noted that the applicant showed extreme insensitivity.
    \vspace{1mm}

    8. On \redactedsmall{} the body declined to accept an appeal submitted by
    the applicant and rejected the applicant's request to review the appeal
    beyond the statutory deadline.
    \vspace{1mm}

    9. Subsequently, the applicant was transferred to the Lublin Remand Centre.
    \vspace{1mm}

    10. On \redactedsmall{} the Lublin Remand Centre Penitentiary Commission
    considered and confirmed the decision to impose the custody conditions for
    the applicant, who was deemed a major threat to the detention facility's safety.
    The applicant appealed.
    \vspace{1mm}

    11. On \redactedsmall{} the Lublin Regional Court dismissed the appeal. The
    court considered the grounds presented in the decision of \redactedsmall{},
    stating that the decision had been lawful and reasonable.
    \vspace{1mm}

    12. On \redactedsmall{} the applicant submitted a complaint to the \redactedsmall{}
    of Prisons. He expressed that he faced challenges in engaging in educational
    programs at the Lublin Remand Centre. The officials rejected the complaint, asserting
    that educational programs were not provided for dangerous prisoners.
    \vspace{1mm}

    13. On an unspecified date in \redactedsmall{} the applicant notified the
    authorities of his allegedly illegitimate designation as a "dangerous individual"
    on \redactedsmall{}. He asserted that the \redactedsmall{} authorities had overstepped
    their responsibilities in applying the regime to him. On \redactedsmall{} the
    \redactedsmall{} refused to initiate an investigation, stating that no violation
    had occurred.
    \vspace{1mm}

    14. On \redactedsmall{} the \redactedsmall{} withdrew its determination to classify
    the applicant as a "dangerous individual." It noted that the applicant's
    conduct had changed and that he no longer presented a risk to the safety of
    the remand center.
    \vspace{1mm}

    %\end{small}
\end{tcolorbox}

Finally, one can push the rewriting even further and also query the LLM to rephrase
all \textit{combinations} of N-grams (up to a maximum arity of 3) that only
occur once or twice in the collection. This rewriting further enhances the robustness
of the text against search-based linkage attacks, although at the cost of more
sweeping changes to the content of the document:

\begin{tcolorbox}
    [enhanced, width=1.0\textwidth, colback=gray!5, colframe=gray!80, boxrule=0.8pt,
    breakable, title=De-identified case $d'$ after rewriting N-gram combinations
    of arity $\in \{2{{,}}3\}$, before skip=10pt, after skip=10pt]
    %\begin{small}

    \textbf{PROCEDURE }
    \vspace{1mm}

    1. The case originated in an application (no. \redactedsmall{}) against
    \redactedsmall{} lodged with the Court under Article 34 of the Convention for
    the Protection of Human Rights and Fundamental Freedoms("the Convention") by
    a \redactedsmall{} national, Mr \redactedsmall{}("the applicant"), on \redactedsmall{}.
    \vspace{1mm}

    2. The applicant, who had been granted legal aid, was represented by
    \redactedsmall{}, a lawyer practising in \redactedsmall{}. The Polish
    Government("the Government") were represented by their Agent, \redactedsmall{},
    of the Ministry of Foreign Affairs.
    \vspace{1mm}

    3. The applicant alleged, in particular, a breach of Article 3 of the
    Convention on account of the imposition on him of the so-called
    \redactedsmall{} regime.
    \vspace{1mm}

    4. On \redactedsmall{} the application was communicated to the Government.
    \vspace{2mm}

    \textbf{THE FACTS }
    \vspace{1mm}

    \textbf{I. THE CIRCUMSTANCES OF THE CASE }
    \vspace{1mm}

    5. The applicant was born in \redactedsmall{} and is detained in \redactedsmall{}.
    \vspace{2mm}

    \textbf{A. Criminal proceedings against the \redactedsmall{} }
    \vspace{1mm}

    6. On \redactedsmall{} the \redactedsmall{} was arrested and remanded in
    custody. In \redactedsmall{} he was convicted of three counts of assault and
    theft and was handed a prison sentence spanning from \redactedsmall{}. On
    \redactedsmall{} the \redactedsmall{} Court handed down a total sentence
    across four offenses.
    \vspace{2mm}

    \textbf{B. Application of the so-called "high-risk detainee" regime }
    \vspace{1mm}

    \redactedsmall{}. On \redactedsmall{} the \redactedsmall{} Court ordered on
    the individual the so-called "high-risk detainee" regime. The body concluded
    that the applicant had been the head of or involved in a coordinated group action
    in \redactedsmall{} (Article \redactedsmall{} § 2 (2a) of the Code of
    Execution of Criminal Sentences), as the individuals declined the meals
    supplied by the prison authorities. The authorities discovered that the individuals
    were organizing a new demonstration for \redactedsmall{}." The body also
    noted that the applicant showed extreme insensitivity.
    \vspace{1mm}

    8. On \redactedsmall{} the body declined to accept an appeal submitted by the
    applicant and rejected the applicant's request to review the appeal beyond the
    statutory deadline.
    \vspace{1mm}

    9. Subsequently, the applicant was transferred to the \redactedsmall{}.
    \vspace{1mm}

    10. On \redactedsmall{} the \redactedsmall{} Commission considered and confirmed
    the decision to impose the custody conditions for the applicant, who was
    deemed a major threat to the detention facility's safety. The applicant appealed.
    \vspace{1mm}

    11. On \redactedsmall{} the \redactedsmall{} Court dismissed the appeal. The
    court considered the grounds presented within the ruling of \redactedsmall{},
    stating that the decision had been lawful and reasonable.
    \vspace{1mm}

    12. On \redactedsmall{} the applicant submitted a complaint to the \redactedsmall{}
    of \redactedsmall{}. He expressed that he faced challenges in engaging in
    education at the \redactedsmall{} Remand Centre. The officials rejected the
    complaint, asserting that education was not provided for high-risk prisoners.
    \vspace{1mm}

    13. On an unspecified date in \redactedsmall{} the applicant notified the
    authorities of his allegedly illegitimate designation as a "dangerous individual"
    on \redactedsmall{}. He asserted that the \redactedsmall{} authorities had overstepped
    their responsibilities in the application of the system against him. On
    \redactedsmall{} the \redactedsmall{} refused to initiate an investigation, stating
    that no violation had occurred.
    \vspace{1mm}

    14. On \redactedsmall{} the \redactedsmall{} withdrew its determination to designate
    the applicant as a "dangerous individual." It noted that the applicant's
    conduct had changed and that he no longer presented a risk to the safety of
    the remand center.

    %\end{small}
\end{tcolorbox}

In this last example, the LLM also replaced some N-grams with \redacted{} when those
were impossible to paraphrase. This is particularly the case for named entities
such as organisations.

\subsubsection{LLM rewriting without specified spans to rephrase}

As discussed in Section \ref{sec:evaluation}, a straightforward baseline
involves prompting the LLM to rewrite the de-identified document without
supplying the specific N-grams to be rephrased. The resulting text was generated
over three rewriting iterations, using the same prompt as outlined in Section
\ref{sec:prompt}, but excluding the span specifications.

\begin{tcolorbox}
    [enhanced, width=1.0\textwidth, colback=gray!5, colframe=gray!80, boxrule=0.8pt,
    breakable, title=De-identified court case $d'$ after LLM rewriting without
    specified spans to rephrase, before skip=10pt, after skip=10pt]
    %\begin{small}

    \textbf{PROCEDURE }
    \vspace{1mm}

    1. The case began with a claim (no. \redactedsmall{}) against \redactedsmall{}
    submitted to the Court under Article 34 of the Convention for the Protection
    of Human Rights and Fundamental Freedoms ("the Convention") by a \redactedsmall{}
    national, Mr \redactedsmall{} ("the applicant"), on \redactedsmall{}.
    \vspace{1mm}

    2. The applicant, who had been granted legal aid, was represented by
    \redactedsmall{}, a lawyer based in Lublin. The Polish Government ("the Government")
    was represented by their lawyer, \redactedsmall{}, of the Ministry of
    Foreign Affairs.
    \vspace{1mm}

    3. The applicant alleged, in particular, a breach of Article 3 of the
    Convention on account of the imposition on him of the so-called "dangerous detainee"
    regime.
    \vspace{1mm}

    4. On \redactedsmall{} the application was communicated to the Government.
    \vspace{2mm}

    \textbf{THE FACTS }
    \vspace{1mm}

    \textbf{I. THE CIRCUMSTANCES OF THE CASE }
    \vspace{1mm}

    5. The applicant was born in \redactedsmall{} and is detained in
    \redactedsmall{}.
    \vspace{2mm}

    \textbf{A. Criminal proceedings against the applicant }
    \vspace{1mm}

    6. On \redactedsmall{} the applicant was arrested and remanded in custody. In
    \redactedsmall{} he was convicted of three counts of battery and robbery and
    received prison terms ranging from \redactedsmall{}. On \redactedsmall{} the
    Lublin Regional Court imposed a cumulative sentence for four criminal convictions.
    \vspace{2mm}

    \textbf{B. Imposition of the so-called "dangerous detainee" regime }
    \vspace{1mm}

    7. On \redactedsmall{} the Opole Lubelskie Prison Penitentiary Commission
    imposed on the applicant the so-called "dangerous detainee" regime. The commission
    determined that the applicant had been the leader or an active participant in
    a planned collective remonstrance in \redactedsmall{} (Article 88a § 2 (2a) of
    the Code of Execution of Criminal Sentences), as the detainees refused the food
    provided by the prison authorities. The authorities learned that the
    detainees were planning another collective protest for \redactedsmall{}. The
    commission also considered that the applicant was particularly egregious.
    \vspace{1mm}

    8. On \redactedsmall{} the \redactedsmall{} rejected an appeal filed by the
    applicant and denied his request to examine the appeal outside the statutory
    time-limit.
    \vspace{1mm}

    9. Subsequently, the applicant was moved to the \redactedsmall{}.
    \vspace{1mm}

    10. On \redactedsmall{} the \redactedsmall{} examined and confirmed the
    decision to apply the regime to the applicant, considering that he posed a serious
    threat to the security of the detention centre. The applicant appealed.
    \vspace{1mm}

    11. On \redactedsmall{} the \redactedsmall{} dismissed the appeal. The court
    cited the grounds outlined in the ruling, holding that the decision was
    lawful and reasonable.
    \vspace{1mm}

    12. On \redactedsmall{} the applicant filed a grievance with the \redactedsmall{}
    of Prisons. He struggled to access learning opportunities at the \redactedsmall{}
    Centre. The authorities dismissed the complaint, stating that learning
    opportunities were not provided for dangerous prisoners.
    \vspace{1mm}

    On an unspecified date in \redactedsmall{} the applicant informed the police
    of his asserted unlawful classification as a "dangerous detainee" on \redactedsmall{}.
    He asserted that the Opole Lubelskie Prison authorities had improperly imposed
    the regime on him. On \redactedsmall{} the Opole Lubelskie District Prosecutor
    refused to initiate an inquiry in the case, holding that no offence had been
    committed.
    \vspace{1mm}

    14. On \redactedsmall{} the \redactedsmall{} revoked the decision to
    classify the applicant as a "dangerous detainee." It determined that the applicant’s
    behaviour had improved and that he no longer posed a risk to the security of
    the remand centre.

    %\end{small}
\end{tcolorbox}

While numerous sentences in the input document have been paraphrased, the rewriting
appears largely indiscriminate -- many frequent N-grams are altered
unnecessarily, whereas the majority of the rare N-grams (22 out of 30) leading to
linkages with the original document $d$ remain unchanged.

%\newpage

\subsubsection{Controlled paraphrasing}

As detailed in Appendix \ref{sec:controlled_paraphrasing}, we also implemented
an paraphrasing approach relying on hancrafted rules. More specifically, the
approach first parsed the document to obtain POS tags and morphological features.
Nouns and verbs were then replaced by synonyms, while heuristic rules were
employed to add or delete modifiers such as adjectives and adverbs.

The outcome of this approach on our example is as follows:

\begin{tcolorbox}
    [enhanced, width=1.0\textwidth, colback=gray!5, colframe=gray!80, boxrule=0.8pt,
    breakable, title=De-identified court case $d'$ after using controlled
    paraphrasing, before skip=10pt, after skip=10pt]
    %\begin{small}

    \textbf{PROCEDURE }
    \vspace{1mm}

    1. The case originated in an application (no. \redactedsmall{}) against
    \redactedsmall{} lodged with the Court under Article 34 of the Convention for
    the Protection of Human Rights and Fundamental Freedoms (“the Convention”) by
    a \redactedsmall{} national, Mr \redactedsmall{} (“the applicant”), on \redactedsmall{}.
    \vspace{1mm}

    2. The applicant, who had been granted legal aid, was represented by
    \redactedsmall{}, a lawyer practising in Lublin. The Polish Government (“the
    Government”) were represented by their Agent, \redactedsmall{}, of the
    Ministry of Foreign Affairs.
    \vspace{1mm}

    3. The applicant alleged, in particular, a breach of Article 3 of the
    Convention on account of the imposition on him of the so-called “life-threatening
    detainee” authorities.
    \vspace{1mm}

    4. On \redactedsmall{} the application was communicated to the Government.
    \vspace{2mm}

    \textbf{THE FACTS }
    \vspace{1mm}

    \textbf{I. THE CIRCUMSTANCES OF THE CASE }
    \vspace{1mm}

    5. The applicant was born in \redactedsmall{} and is detained in
    \redactedsmall{}.
    \vspace{2mm}

    \textbf{A. Criminal proceedings against the applicant }
    \vspace{1mm}

    6. On \redactedsmall{} the applicant was arrested and remanded in custody. In
    \redactedsmall{} he was convicted of three counts of bombardment and looting
    and received prison house sentences ranging from \redactedsmall{}. On
    \redactedsmall{} the lublin court imposed a cumulative time for four felonious
    convictions.
    \vspace{2mm}

    \textbf{B. infliction of the so-called “ unsafe political detainee ” authorities
    }
    \vspace{1mm}

    7. On \redactedsmall{} the Opole Lubelskie Prison Penitentiary Commission
    imposed on the applicant the so-called “ unsafe detainee” regime. the direction
    held that the applier had been the leader of or an participating player in a
    planned remonstration in \redactedsmall{} (Article 88 amp § 2 (2a) of the Code
    of Execution of Criminal Sentences), as the detainees refused to have food for
    thought provided by the prison house authorities. The authorities learned
    that the detainees were planning another protest for \redactedsmall{}. The
    delegation also considered that the applicant was in particular perverted.
    \vspace{1mm}

    8. On \redactedsmall{} the Opole Lubelskie Prison Penitentiary Commission rejected
    an appeal lodged by the applier and dismissed his petition for leave of absence
    to prove the solicitation outside the statutory time-limit.
    \vspace{1mm}

    9. Subsequently, the applicant was transferred to the Lublin Remand Centre.
    \vspace{1mm}

    10. On \redactedsmall{} the Lublin Remand Centre Penitentiary Commission reviewed
    and upheld the decisiveness to implement the authorities to the applicant,
    considering that he posed a sober threat to the protection of the detention
    centre. The applicant appealed.
    \vspace{1mm}

    11. On \redactedsmall{} the lublin Court dismissed the appeal. The judicature
    referred to the reasons given in the decisiveness of \redactedsmall{}, holding
    that the decision had been lawful and justified.
    \vspace{1mm}

    12. On \redactedsmall{} the applicant lodged a complaint with the lublin
    Inspectorate of Prisons. He complained that he was having difficulties in accessing
    activities in the Lublin Remand Centre. The authorities dismissed the
    complaint, holding that activities were not unionized for dangerous prisoners.
    \vspace{1mm}

    13. On an unspecified date in \redactedsmall{} the applicant notified the
    police of his wrongful assortment as a “ unsafe political detainee ” on
    \redactedsmall{}. He claimed that the Opole Lubelskie Prison authorities had
    exceeded their powers in imposing the regimen on him. On \redactedsmall{}
    the Opole Lubelskie District Prosecutor refused to open an investigation in
    the case, holding that no discourtesy had been committed.
    \vspace{1mm}

    14. On \redactedsmall{} the Lublin Remand Centre Penitentiary Commission
    revoked its decision to sort out the applicant as a “life-threatening
    detainee”. it considered that the applicant’s demeanor had improved and that
    he no longer posed a threat to the security of the remand centre.

    %\end{small}
\end{tcolorbox}

We can observe that the approach is able to maintain the semantic content of the
text, although a substantial number of linkable N-grams remain present. In addition,
the lack of context-sensitivity of the rewriting mechanism does sometimes lead
to suboptimal or wrong replacements, such as "battery and robbery" being replaced
by "bombardment and looting".

%\newpage

\subsubsection{DP-based rewriting}

One final alternative is to rely on approaches grounded in Differential Privacy (DP).
We have notably experimented with the DPMLM model of \cite{Meisenbacher2024-ib}.
Unfortunately, the quality of the resulting documents is very poor (here with an
$\epsilon = 100$):

\begin{tcolorbox}
    [enhanced, width=1.0\textwidth, colback=gray!5, colframe=gray!80, boxrule=0.8pt,
    breakable, title=De-identified court case $d'$ after DP-based rewriting with
    $\epsilon{{=}}100$, before skip=10pt, after skip=10pt]
    %\begin{small}

    State 5. The cases originated in an application (no. [Red]) against [Red] lodged
    with the Judge under Article 25 of the Conference for the Conservation of Human
    Right and Human Issues (" the Convention () by a [Acted] national, Mr [Red]
    (" the applicant (), on [Red].
    \vspace{1mm}

    3. This applicant, who had been received fraudulent aiding, was testified by
    [Red], a comedian pract in Warsaw. That Greek State (" the Governments?)
    were featured by their Agents, [Acted], of the Government of Foreign Arts.
    \vspace{1mm}

    4. They application did, in particulars, a break of Articles iii of the
    Convention on aim of the implementation on him of the the " dangerous
    detainee regime.
    \vspace{1mm}

    4. In [Red] the application was given to the Government. Related Facts O.
    The Background Are Your Court
    \vspace{1mm}

    8. The petitioner was born in [Red] and is held in [Red]. A. Criminal cases against
    the appellant
    \vspace{1mm}

    6. Off [Red] the respondent was handcuffed and put in custody. In [Red] he was
    condemned of 43 monsters of battery and robbers and earned incarcerated orders
    scoring from [Red]. After [Red] the Moscow R Department announced a
    cumulative suspension for four different punishments.
    \vspace{1mm}

    6. Introduction of the alleged " dangerous detainee dictatorship
    \vspace{1mm}

    9. On [Red] the Op K Prison Pen Commission initiated on the applicant the " dangerous
    detainee dictatorship. The commission withheld that the applicants had been
    the leader of or an advanced plaintiff in a packed collectively
    demonstration in [Redacted] (Articles 88 § 3 (2) of the Title of Administration
    of Criminal Ences), as the patients determined to permit material released by
    the jail complexes. Some courts suspected that the occupants were practicing
    a collectively confrontation for [Red]. This commissioner apparently deemed that
    the respondent was additionally sexist.
    \vspace{1mm}

    3. And [Red] the Ole C Province Prison Commission revoked an intervention received
    by the holder and released his attempt for permission to view the appeal after
    the contractual limitation.
    \vspace{1mm}

    14. Upon, the applicant was copied to the Munich Reform Center.
    \vspace{1mm}

    14. And [Red] the Munich Detention Centre Prison Commission appealed and
    approved the proposal to grant the regime to the appellant, forwarding that he
    issued a deep menace to the health of the incarceration room. They holder
    appeals.
    \vspace{1mm}

    11. On [Red] the Warsaw Superior Cour revived the argument. The court referring
    to the examples passed in the declaration of [Red], holding that the case
    had been just and decisive.
    \vspace{1mm}

    3. After [Acted] the application directed a complain with the Berlin Rural
    Inventory of Isons. Once said that he was having supports in attending educational
    facilities in the Berlin Detention Centre. State authorities closed the
    matter, hold that instructional events were not meant for serious ones.
    \vspace{1mm}

    28. Before an mentioned date in [Red] the applicants briefed the police of his
    unbelievably lawful listing as a « dangerous inmate on [Acted]. Also
    suggested that the Op Kie Prison apparatus had unleashed their policies in
    assuming the regime on him. When [Acted] the Warsaw Cki Public Prosecutor ruled
    to launch an exhibition in the cases, restraining that no offence had been
    confessed.
    \vspace{1mm}

    19. : [Red] the Berlin And Centre Correctional Committee repealed its
    decision to exclude the applicants as a " dangerous offender ". In reached that
    the attacker? s act had changed and that he no again ranked a challenge to
    the quality of the and centre.

    %\end{small}
\end{tcolorbox}

%\subsection{Algorithm}

%\begin{algorithm}
%\caption{Find Linked N-Grams in a De-identified Document}
%\begin{algorithmic}[1]
%\Require Inverted index $\mathcal{I}$ including the N-grams occurring in $\mathcal{D}$
%\Require De-identified document $d'$
%\Require Minimum threshold $k$ on number of documents
%\Procedure{FindLinkedNGrams}{$d', \mathcal{I}$}
%    \State Extract N-grams: $S_{d'} \gets \{s \mid s \text{ is a contiguous sequence of tokens in } d'\}$
%    \State Sort substrings in $S_{d'}$ by their length
%   \State Remove overlapping substrings from $S_{d'}$
%    \For{$\text{comb\_length} \in [1, \text{max arity}]$}
%        \State Generate N-gram combinations: $C \gets C(S_{d'}, \text{comb\_length})$
%        \For{$\text{comb} \in C$}
%            \State Get posting lists: $P \gets \{\mathcal{I}(s) \mid s \in \text{comb}\}$
%            \State Intersect posting lists: $\text{docs} \gets \text{intersect}(P)$
%            \If{$|\text{docs}| < k$}
%                \State \textbf{yield} comb
%            \EndIf
%        \EndFor
%    \EndFor
%\EndProcedure
%\end{algorithmic}
%\caption{}
%\end{algorithm}
\end{document}